\documentclass[final]{cvpr}

\usepackage{times}
\usepackage{amsmath,amssymb,graphicx}
\usepackage{stmaryrd}
\usepackage{amsfonts}
\usepackage{url}
\usepackage{algorithm}
\usepackage{algorithmicx}
\usepackage{algpseudocode}
\usepackage{booktabs}
\usepackage{adjustbox}
\usepackage{threeparttable}
\usepackage{multirow}
\usepackage{subcaption}
\usepackage{caption}
\usepackage{array}
\usepackage{color}

\DeclareMathOperator*{\argmax}{argmax}

\graphicspath{{figure/}}

\def\blue#1{\textcolor[rgb]{0,0,1}{#1}}
\def\red#1{\textcolor[rgb]{1,0,0}{#1}}

\def\fred#1{\textcolor[rgb]{0,0,0}{#1}}

\newcommand{\keypoint}[1]{\vspace{0.05cm}\noindent\textbf{#1}\quad}
\usepackage{ragged2e}
\renewcommand{\raggedright}{\leftskip=0pt \rightskip=0pt plus 0cm}

\usepackage[pagebackref=true,breaklinks=true,letterpaper=true,colorlinks,bookmarks=false]{hyperref}

\begin{document}

\title{Your ``Flamingo'' is My ``Bird'': Fine-Grained, or Not}

\author{Dongliang~Chang,~Kaiyue~Pang,~Yixiao~Zheng,~Zhanyu~Ma,~Yi-Zhe~Song, and~Jun~Guo}  

\maketitle

\begin{abstract}
Whether what you see in Figure 1 is a ``\fred{flamingo}'' or a ``\fred{bird}'', is the question we ask in this paper. While fine-grained visual classification (FGVC) strives to arrive at the former, for the majority of us non-experts just ``\fred{bird}'' would probably suffice. The real question is therefore -- how can we tailor for different fine-grained definitions under divergent levels of expertise. For that, we re-envisage the traditional setting of FGVC, from single-label classification, to that of top-down traversal of a pre-defined coarse-to-fine label hierarchy -- so that our answer becomes ``\fred{bird}'' 
$\Rightarrow$ 
``\fred{Phoenicopteriformes}'' 
$\Rightarrow$ 
``\fred{Phoenicopteridae}''
$\Rightarrow$
``\fred{flamingo}''. 

To approach this new problem, we first conduct a comprehensive human study where we confirm that most participants prefer multi-granularity labels, regardless whether they consider themselves experts. We then discover the key intuition that: coarse-level label prediction exacerbates fine-grained feature learning, yet fine-level feature betters the learning of coarse-level classifier. This discovery enables us to design a very simple albeit surprisingly effective solution to our new problem, where we (i) leverage level-specific classification heads to disentangle coarse-level features with fine-grained ones, and (ii) allow finer-grained features to participate in coarser-grained label predictions, which in turn helps with better disentanglement. Experiments show that our method achieves superior performance in the new FGVC setting, and performs better than state-of-the-art on the traditional single-label FGVC problem as well. Thanks to its simplicity, our method can be easily implemented on top of any existing FGVC frameworks and is parameter-free. 

\end{abstract}

\footnote{D. Chang, Y. Zheng, Z. Ma, and J. Guo are with the Pattern Recognition and Intelligent
System Laboratory, School of Artificial Intelligence, Beijing University of Posts and Telecommunications, Beijing 100876, China (e-mail: mazhanyu@bupt.edu.cn).}
\footnote{K. Pang and Y.-Z. Song  are with SketchX, CVSSP, University of Surrey, London, United Kingdom.}

\vspace{-10mm}
\section{Introduction}

\begin{figure}[t]
\begin{center}
  \includegraphics[width=1.0\linewidth]{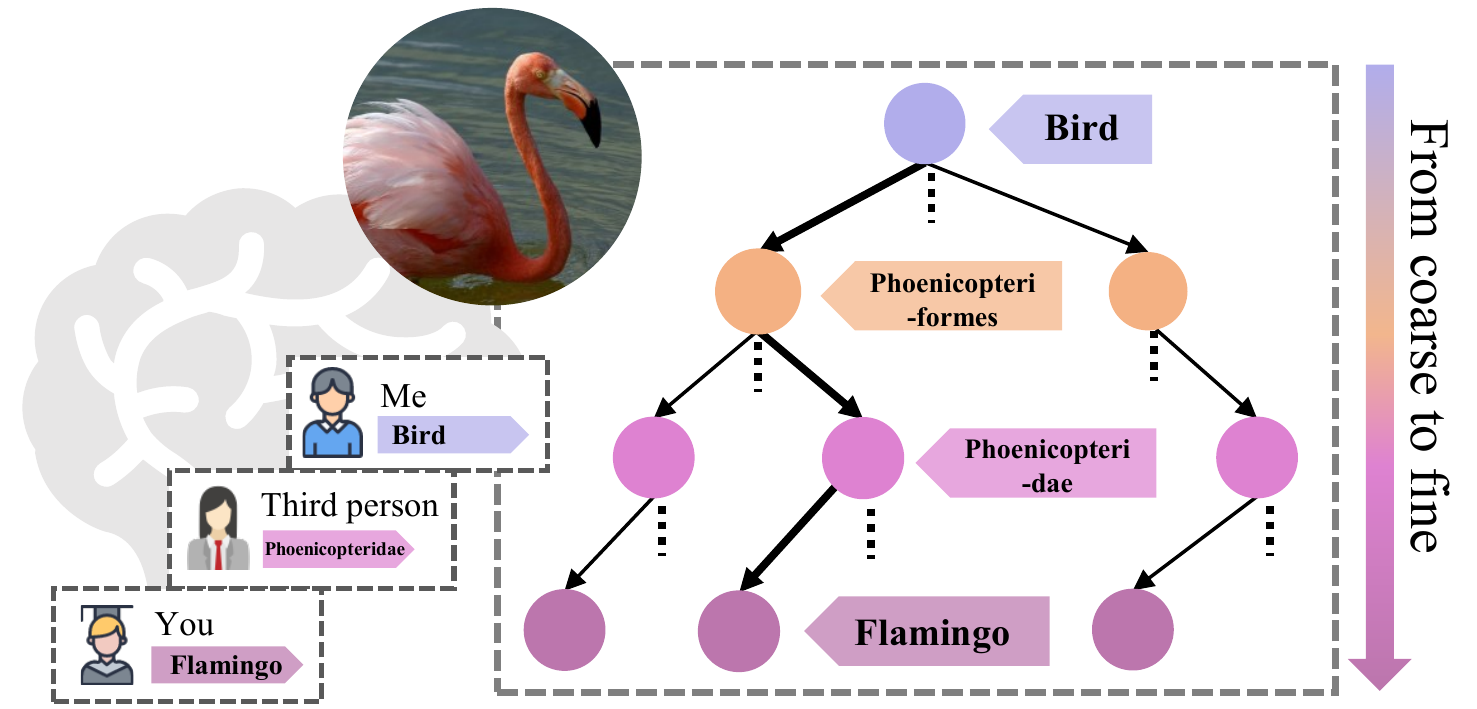}
\end{center}

  \caption{Definition of what is fine-grained is subjective. Your ``\fred{flamingo}'' is my ``\fred{bird}''.}
\label{fig:Labrador}
\vspace{-5mm}
\end{figure}
Fine-grained visual classification (FGVC) was first introduced to the vision community almost two decades ago with the landmark paper of \cite{biederman1999subordinate}. It brought out a critical question that was largely overlooked back then -- that can machines match up to humans on recognising objects at fine-grained level (e.g., a ``\fred{flamingo}'' other than a ``\fred{bird}''). Great strides have been made over the years, starting with the conventional part-based models \cite{yao2012codebook,gao2013learning,berg2013poof,branson2014bird}, to the recent surge of deep models that either explicitly or implicitly tackle part learning with or without strong supervision \cite{lin2015bilinear,Peng2018Object,zhang2016picking,zheng2017learning,Zheng_2019_CVPR,wu2019leveraging}. Without exception, the focus has been on mining fine-grained discriminative features to better classification performances.

In this paper, we too are interested in the fine-grained rationale at large -- yet we do not set out to pursue performance gains,  we instead question the very definition of fine-grained classification itself. In particular, we ask whether the fine-grained expert labels commonplace to current FGVC datasets indeed convey what end users are accustomed to -- i.e., are the ``Florida scrub jay'', ``Fisker Karma Sedan 2012", ``Boeing 737-200" are indeed the desired, or would  ``\fred{bird}'', ``car", ``aircraft" suffice for many -- my ``\fred{flamingo}'' can be just your ``\fred{bird}''. The answer is of course subjective~\cite{ordonez2015predicting}, and largely correlates with expert knowledge -- the more you are a \fred{bird} lover, the more fine-grained labels you desire, some might even look for ``\fred{American flamingo}'' other than just ``\fred{flamingo}''. The follow-up question is therefore, how can we tailor for the various subjective definitions of what is fine-grained, and design a system that best accommodates practical usage scenarios of FGVC. 

To answer this, we first conduct a human study on the popular CUB-200-2011 bird dataset \cite{wah2011caltech} with two questions in mind (i) how useful are the pre-defined fine-grained labels to a general user, and (ii) whether a single label output is in fact a preferred solution. We first build a hierarchical taxonomy of bird, by tracing existing fine-grained labels in CUB-$200$-$2011$ to its parent sub-category, all the way to the super node of ``bird'' using Wikipedia.  We then recruited $50$ participants with various background of bird knowledge, each of whom rated $100$ bird photos by (i) picking a label amongst fine- and coarse-grained ones relating to the bird, and (ii) indicating whether more label choices are desirable other than just the single label previously selected. We find that (i) participants do not necessarily choose the pre-defined fine-grained (bottom) labels as their preferred choice, (ii) only $36.4\%$ of all returned choices prefer just a single label, and (iii) although domain experts tend to choose finer-grained labels while amateurs prefer coarser ones, close to $80\%$ of choices from experts also turn to the option of multi-granularity labels.

Following results from the human study, we propose to re-instantiate the FGVC problem by extending it from a \textit{single-label classification} problem, to that of \textit{multiple label predictions} on a pre-defined label hierarchy. The central idea is while people tend to feel baffled facing a single expert label, a chain of coarse-to-fine labels that describe an object can potentially be more practical -- we leave it to the users to decide which fine-grained level along the hierarchy best suits their needs. Compared with a single label telling you it is a ``\fred{flamingo}'' (as per conventional FGVC), our model offers a coarse-to-fine series of labels such as 
``\fred{bird}'' 
$\Rightarrow$ 
``\fred{Phoenicopteriformes}'' 
$\Rightarrow$ 
``\fred{Phoenicopteridae}''
$\Rightarrow$
``\fred{flamingo}'' (See Figure \ref{fig:Labrador}).

On the outset, classifying an image into multiple cross-granularity labels seems an easy enough extension to the well-studied problem of FGVC with single-label output. One can simply train a single model for classifying all nodes in the hierarchy, or better yet use separate classifiers for each hierarchy level. Although these do work as baselines, they do not benefit from the inherent coarse-fine hierarchical relationship amongst labels -- we show exploring these relationships not only helps to solve for the new FGVC setting, but also in turn benefits the learning of fine-grained features which then helps the conventional task.

Our design is based on the discovery of two key observations on the label hierarchy: (i) coarse-level features in fact exacerbates the learning of fine-grained features, and (ii) finer-grained label learning can be exploited to enhance the discriminability of coarser-grained label classifier. Our first technical contribution is therefore a multi-task learning framework to perform level-wise feature disentanglement, with the aim to separate the adverse effect of coarse feature from fine-grained ones. To further encourage the disentanglement, we then resort to the clever use of gradients to reflect our second observation. Specifically, during the forward pass only, we ask finer-grained features to participate in the classification of coarser-grained labels via feature concatenation. We, however, constrain the gradient flow to only update the parameters within each multi-task head. Our method is generic to any existing FGVC works and experiments show that it yields stronger classifiers across all granularities. Interestingly, our model also delivers state-of-the-art result when evaluated on the traditional FGVC setting, while not introducing any additional parameters. 

Our contributions are as follows: (i) we re-envisage the problem setting of FGVC, to accommodate the various subjective definitions of ``fine-grained'', where we advocate for top-bottom traversal of a coarse-to-fine label hierarchy, other than the traditional single-label classification; (ii) we discover important insights on the inherent coarse-fine hierarchical relationship to drive our model design, and (iii) we show by disentangling coarse-level feature learning with that of fine-grained, state-of-the-art performances can be achieved both on our new problem, and on the traditional problem of FGVC.

\begin{figure*}[!t]
  \centering
   \begin{subfigure}[t]{2.252in}
    \centering
    \includegraphics[width=1.802in]{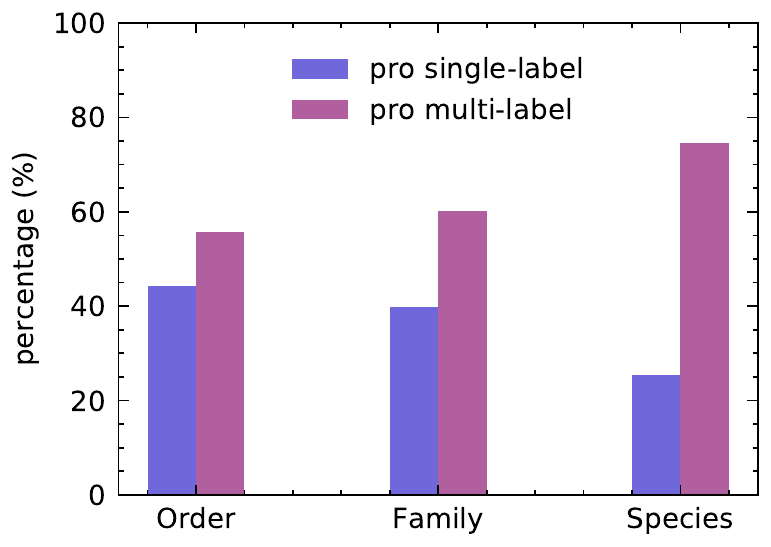}
    \caption{}
 \end{subfigure}
  \begin{subfigure}[t]{2.265in}
    \centering
    \includegraphics[width=1.805in]{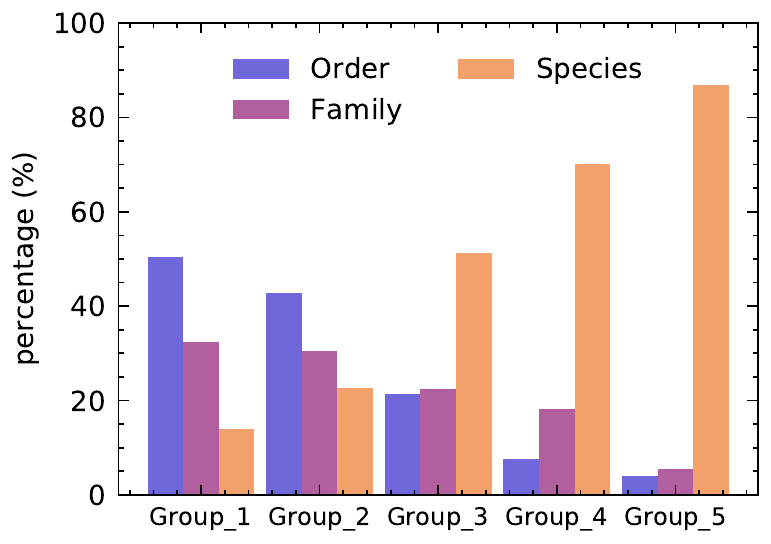}
    \caption{}
 \end{subfigure}
   \begin{subfigure}[t]{2.265in}
    \centering
    \includegraphics[width=1.805in]{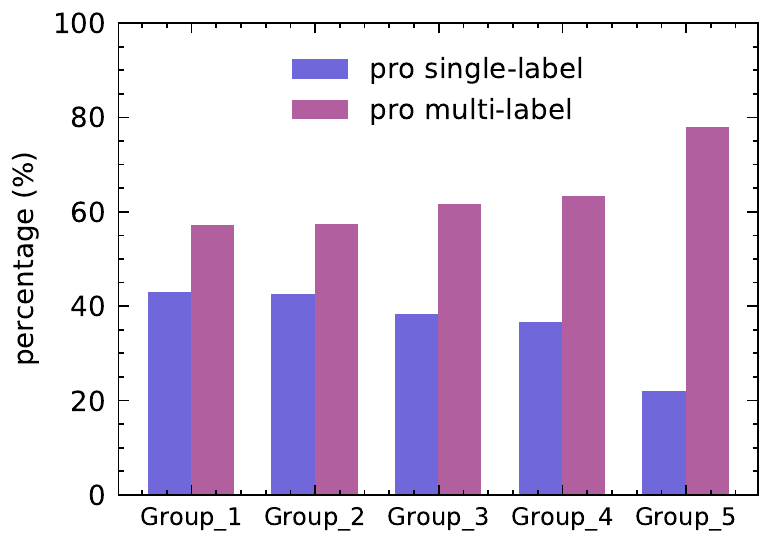}
    \caption{}
 \end{subfigure}

  \caption{Human study on CUB-200-2011 bird dataset. Order, family, species are three coarse-to-fine label hierarchy for a bird image. A higher group\_id represents a group of people with better domain knowledge of birds, with group\_5 interpreted as domain experts. (a) Human preference between single and multiple labels. (b) Impact of human familiarity with birds on single-label choice. (c) Impact of human familiarity with birds on multi-label choice.}
  \label{fig:Human_study}

\end{figure*}

\vspace{-2mm}
\section{Related Work}

\keypoint{Fine-grained image classification} Deep learning has emerged as powerful tool that led to remarkable breakthroughs in FGVC~\cite{zhang2016weakly,yang2018learning,wei2019adversarial,lu2019using,chen2019destruction,zhuang2020learning,ji2020attention}. Compared with generic image recognition task~\cite{cornia2020meshed,wang2020dual}, FGVC requires a model to pay special attention on the very subtle and local image regions~\cite{yang2018learning,chang2020mc}, which are usually hard to notice in human eyes. 
A major stream of FGVC works thus undergoes two stages by first adopting a localisation subnetwork to localise key visual cues and then a classification subnetwork to perform label prediction. Earlier works on localisation module rely heavily on additional dense part/bounding box annotations to perform detection~\cite{berg2013poof, chai2013symbiotic}, and gradually move towards weakly supervised setting that only requires image labels~\cite{yang2018learning,chang2020mc}. 
Relevant techniques including unsupervised detection/segmentation, utilisation of deep filters and attention mechanism have been proposed to guide the extraction of the most discriminative image regions~\cite{xiao2015application,wang2018learning,huang2020interpretable}. Another line of FGVC research focuses on end-to-end feature encoding~\cite{dubey2018pairwise,sun2018multi,sun2019fine}. This saves the effort of explicit image localisation but asks for extra effort to encourage feature discriminability, e.g., high-order feature interactions~\cite{lin2015bilinear,zheng2019learning}. In this paper, we study a different setting for FGVC that generates multiple output labels at different granularities for an image. 

\keypoint{Multi-task learning} Multi-task learning (MTL) aims to leverage the common information among tasks to improve the generalisability of the model~\cite{chen2018driving,chen2019progressive,luo2020multi,zhou2020pattern}. 
Under the context of deep learning, MTL translates to designing and optimising networks that encourage shared representations under multi-task supervisory signals. There are two types of parameter sharing. The hard way is to divide the parameter set into shared and task-specific operators~\cite{kokkinos2017ubernet, kendall2018multi, chen2018gradnorm}. In soft parameter sharing, however, each task is assigned its own set of parameters and 
further regularisation technique are introduced to encourage cross-task talk ~\cite{misra2016cross,ruder2019latent,gao2019nddr}.
Joint learning of multiple tasks is prone to negative transfer if the task dictionary contains unrelated tasks~\cite{kokkinos2017ubernet,he2017mask}. This problem triggers another line of MTL research with numerous solutions proposed, including reweighing the individual task loss~\cite{kendall2018multi,sinha2018gradient}, 
tailoring task-specific gradient magnitudes~\cite{chen2018gradnorm} and disentangling features between irrelevant tasks~\cite{guo2018dynamic,zhao2018modulation}. 
We approach the multi-task learning in FGVC following a similar underlying motivation - by identifying impacts of transfer between label predictions at different granularities. More specifically, we propose a novel solution to simultaneously reinforce positive and mitigate negative task transfer.

\section{Human Study}\label{Pilot_Study}

To inform the practical necessity of our multiple cross-granularity label setting, we conduct a human study~\cite{granqvist2013hedging} on the CUB-200-2011 bird dataset. This is in order to show (i) single fine-grained label generated by existing FGVC models does not meet the varying subjective requirements for label granularity in practice; (ii) multiple label outputs covering a range of granularity are able to bridge the perceived gaps amongst different populations.

\begin{table}[!t]
  \centering

 \begin{adjustbox}{width=0.9\linewidth,center}
       \footnotesize   
    \begin{tabular}{l|cccc}
    \toprule
    \textbf{Choice} \ \ \ & \textbf{Order} & \textbf{Family} & \textbf{Species} & \textbf{None}\\
    \midrule
    \textbf{Percentage} \ \ \ & $29.2\%$ & $30.4\%$ & $36.4\%$ & $4\%$ \\
    \bottomrule
    \end{tabular}%
 \end{adjustbox}
  \caption{Human preference between labels at different granularity on CUB-200-2011 bird dataset.}
  \label{tab:addlabel}%
\vspace{-4mm}
\end{table}%

\keypoint{Data \& Participant Setup} CUB-200-2011 is a bird dataset commonly used by the FGVC community. It contains $11, 877$ images each labelled as a fine-grained bird species by the domain expert. We extend it by adding two new  hierarchy levels  on top of the species with reference to Wikipedia pages, i.e., identifying the \textit{family} and \textit{order} name for a bird image. This makes each image annotated with three labels at different granularity, in an increasing fineness level from order to species. We performed an initial test amongst $200$ participants across different ages, genders and education levels, to find out their familiarity with birds. We discover that there exists a considerable ``long tail'' problem in their distribution of scores -- there are naturally less bird experts. This motivates us to manually filter for a population that serves as a better basis for statistical analysis. We therefore sample $50$ participants from the original $200$ and encourage the distribution of their expertise (scores) to follow a Gaussian-like shape. We then divide them into $5$ groups ([group\_1, group\_2, ..., group\_5]) based on their scores, where a higher group id corresponds to a population of better domain knowledge. These $50$ participants are included for the task below.

\begin{figure*}[t]
\begin{center}
  \includegraphics[width=0.9\linewidth]{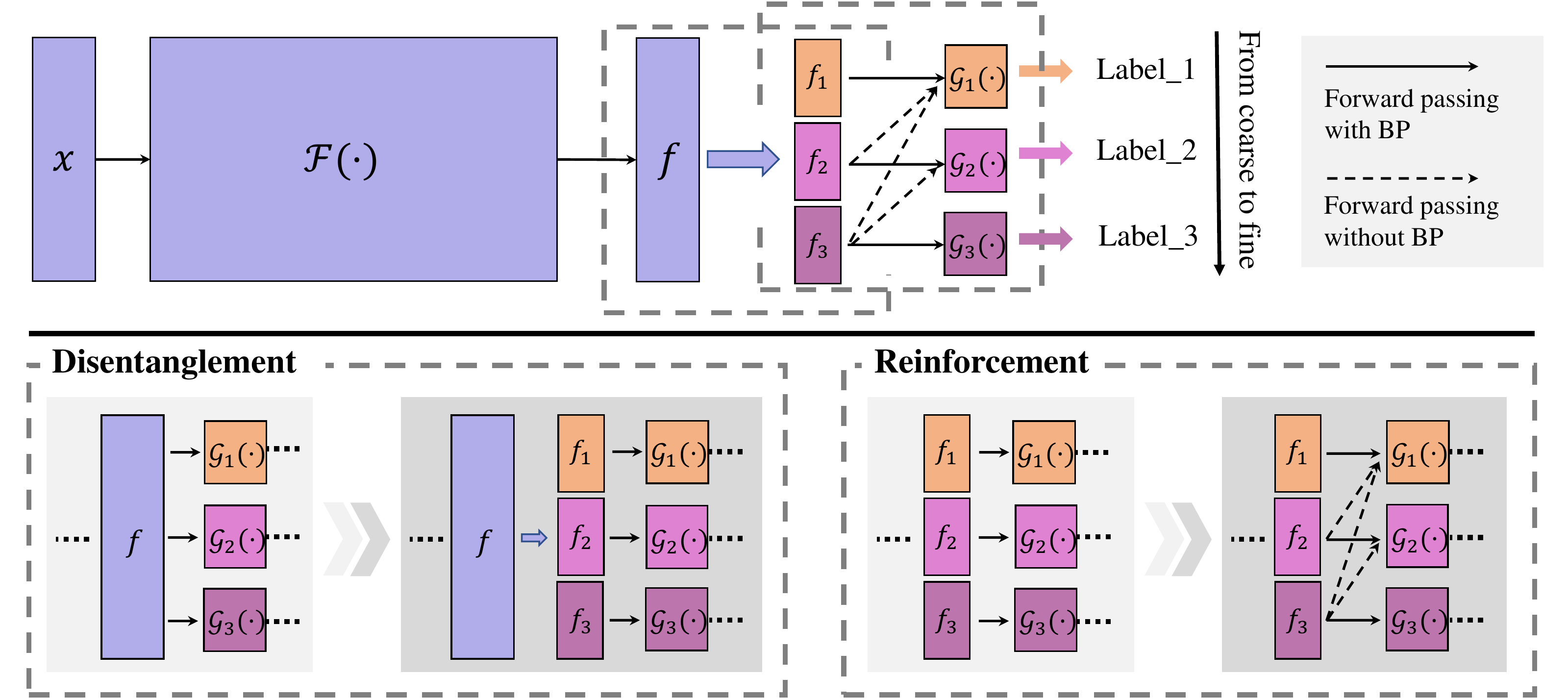}
\end{center}

  \caption{A schematic illustration of our FGVC model with multi-granularity label output.  BP: backpropagation. }

\label{fig:Method}
\end{figure*}

\keypoint{Experiment setting} Designing experiments to validate people's preference on one single label across all granularities is straightforward. But it requires extra consideration for making comparative choices between single and multiple labels. For example, it would not be ideal if we show participants an image with two options of single and multiple labels, since people are naturally biased towards multiple labels as they contain more information ~\cite{thabane2010tutorial}. We therefore design a two-stage experiment, with both stages showing a participant the same image but with different questions.

\noindent\textbf{Stage 1:} \textit{This is a bird. Which one of the labels further defines this bird? You can only choose one option. [A] order\_name [B] family\_name [C] species\_name [D] none of above}

\noindent\textbf{Stage 2:} \textit{At stage 1, do you have the impulse to choose more than one label? [A] yes [B] no}

\noindent Note that participants selecting option D in stage 1 will be directly guided to the next image, skipping stage 2 all together.

\keypoint{Results} We select $1000$ images from CUB-200-2011 and from which, a set of random $100$ images is assigned to each participant. Images received less then three responses are excluded for statistical significance. We analyse the results as follows:

\keypoint{\textit{Your label is not mine}} Table \ref{tab:addlabel} shows the percentage of each option being selected in Stage 1. We can see that (i) participants have varying demands for label granularity; and (ii) The single fine-grained labels (Species option) optimised by existing FGVC models only constitute $36.4\%$ of participant choices in our experiment, while leaving the rest $59.6\%$ (order + family) potentially catered for under a multi-label setting.

\keypoint{\textit{Multiple labels work}} In Figure \ref{fig:Human_study}(a), we show the distribution of preference between single and multiple labels in the second stage. It can be seen that no matter what label (excluding ``None") is chosen in the first stage, the majority of participants turn to embrace multiple labels. This is especially true for participants once selecting species as their single choice, who are the target audience under traditional FGVC setting, and yet still consider multiple cross-granularity labels a better way to interpret an image. 

\keypoint{\textit{Further analysis}} Figure \ref{fig:Human_study}(b) and (c) further show how populations with different familiarity levels with birds lead to different choices in stage 1 and stage 2 respectively. We can see that (i) participants with more domain knowledge (e.g., group\_4) tend to choose finer-grained single labels while amateurs (e.g., group\_1) prefer more interpretable coarser-grained counterparts; (ii) choices under multiple labels have greatly converged regardless of the gaps of domain knowledge. In summary, it is hard to have one level of label granularity that caters to every participant. Multiple cross-granularity labels, however, are found to be meaningful to the many.

\begin{figure*}[t]
\begin{center}
  \includegraphics[width=0.98\linewidth]{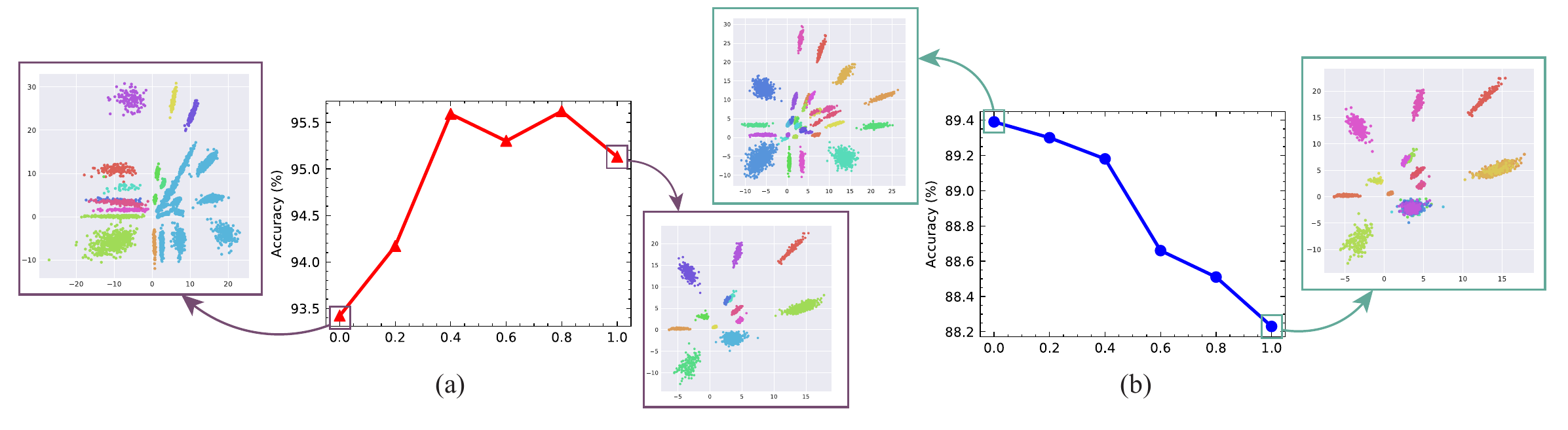}
\end{center}

  \caption{Joint learning of two-granularity labels under different weighting strategy on CUB-200-2011 bird dataset. (a) x-axis: $\beta$ value that controls the relative importance of a fine-grained classifier; y axis: performance of the coarse-grained classifier. (b) x-axis: $\alpha$ value that controls the relative importance of a coarse-grained classifier; y axis: performance of the fine-grained classifier.}

\label{fig:toy_game}
\end{figure*}

\section{Methodology}

Conclusions from our human study motivate us to go beyond the single label output as found in most existing FGVC works, and move towards generating multi-granularity labels. This makes our new setting fall naturally under the multi-task learning framework. Our first goal is to investigate the impact of transfer between label prediction tasks at different granularities. We next build on the insight gained and propose a simple but effective solution that improves the accuracy of label prediction at all granularities. A schematic illustration of our model is shown in Figure \ref{fig:Method}.

\keypoint{Definition} Suppose for each image $x$, we have one fine-grained label $y^K$ from the existing FGVC dataset. To tailor it for our new FGVC setting, we build upon $y^K$ to form $(K-1)$ label hierarchies by finding its superclasses in the Wikipedia pages. This gives us a re-purposed dataset where each image $x$ is annotated with a chain of $K$ labels defined across different granularities, $y^1, y^2, ..., y^k, ...,y^K$. We denote the number of categories within each label granularity as $C_1, C_2, ..., C_k, ... , C_K$, so that $y^k$ is a one-hot vector of length $C_k$. Given any CNN-based network backbone $\mathcal{F}(\cdot)$, We feed $x$ as input to extract its feature embedding {$f=\mathcal{F}(x)$}. Our goal is then to correctly predict labels across $K$ independent classifiers, $\mathcal{G}_1(\cdot), \mathcal{G}_2(\cdot), ..., \mathcal{G}_k(\cdot), ..., \mathcal{G}_K(\cdot)$ based on $f$, i.e., $\hat{y}^k = y^k$, where $\hat{y}^k=\mathcal{G}_k(f)$. Our optimisation objective is $K$ independent cross-entropy loss $\sum_{k=1}^{K}L_{CE}(\hat{y}^k, y^{k})$, and during inference, we take the maximum output probability from each classifier as its label, $l^k=\argmax\limits_{C_{\fred{k}}}\hat{y}^k$.


\subsection{Cooperation or Confrontation?}\label{Cooperation_or_Confrontation}

To explore the transfer effect in the joint learning of multi-granularity labels, we design an image classification task for predicting two labels at different granularities, i.e., $K=2$. We form our train/test set from CUB-$200$-$2011$ bird dataset and assign each image with two labels at order and family level. During training, we introduce two weights as hyperparameters to control the relative importance of each task. This is formulated as:
\begin{equation}
\alpha L_{CE}(\hat{y}^1, y^{1})+\beta L_{CE}(\hat{y}^2, y^{2}) 
\end{equation}

\noindent where a larger value of $\alpha$ and $\beta$ then prioritise feature learning towards predicting coarse-grained and fine-trained labels respectively.

Figure~\ref{fig:toy_game}(a) shows that by keeping $\alpha=1.0$ and gradually increasing the value of $\beta$ from $0.0$ to $1.0$, coarse-grained classifier is constantly reinforced when the features is optimised towards fineness. This is in a stark contrast with Figure~\ref{fig:toy_game}(b) where the performance of fine-grained classifier becomes consistently worse with the increasing proportions of coarse-level features. This provides compelling evidence to the discovery we mentioned earlier: coarse-level label prediction in fact hurts fine-grained feature learning, yet fine-level feature betters the learning of coarse-level classifiers. Such finding is also intuitively understandable because models optimised towards finer-grained recognition are forced to interpret and analyse more local and subtle discriminative regions. They thus comprise additional useful information for coarse-grained classifiers as well. In comparison, features optimised for predicting coarse-grained labels are less likely to generalise.

To provide further proof, we visualise the feature embeddings learned under four weighting strategies using t-SNE, i.e., \{$\alpha=1, \beta=0$\}, \{$\alpha=1, \beta=1$\}, \{$\alpha=0, \beta=1$\}, \{$\alpha=1, \beta=1$\}. Same conclusions still hold. The decision boundaries for coarse-grained label classifiers become more separated with the help of finer-grained features, while fine-grained classifiers are getting worse in this sense given the increasing involvement of coarser-grained features.

\subsection{Disentanglement and Reinforcement}

Observations in Section \ref{Cooperation_or_Confrontation}  suggests that there involves both positive and negative task transfer in multi-granularity label predictions. This leads to our two technical considerations: (i) To restrain from the negative transfer between label predictions at different granularity,  we first explicitly disentangle the decision space by constructing granularity-specific classification heads. (ii) We then implement the potential of positive transfer by allowing fine-grained features to participate in the coarse-grained label predictions and make smart use of gradients to enable better disentanglement.

Specifically, We first split $f$ into $K$ equal parts, with each representing a feature $f_k$ independently responsible for one classifier $\mathcal{G}_k(\cdot)$. To allow finer-grained features in jointly predicting a coarse-grained label $y^k$, we concatenate feature $f_k$ and all the other finer features $f_{k+1},f_{k+2}$,...,${f_K}$ as input to the classifier $\mathcal{G}_k(\cdot)$. One issue remains unsolved. While we have adopted finer-grained features to improve coarse-grained label predictions, this risks the fact that features belonging to fine-grained classifiers will be biased towards coarse-grained recognition during model optimisation and undermines our efforts on disentanglement. We therefore introduce a gradient controller $\Gamma(\cdot)$. That is during the model backward passing stage, we only propagate the gradients flow of one classifier along its own feature dimensions and stop other gradients via $\Gamma(\cdot)$. This gives us final representation of predicting a label:
\begin{equation}
\hat{y}^k = \mathcal{G}_k(CONCAT(f_k, \Gamma(f_k+1), ..., \Gamma(f_K)))
\end{equation}

\begin{table*}[htbp]
  \centering
  \begin{adjustbox}{width=\linewidth,center}
    \Huge   
    \begin{tabular}{l|cccc|cccc|ccc}
    \toprule
    \multirow{2}[4]{*}{\textbf{Method}} & \multicolumn{4}{c|}{\textbf{CUB-200-2011}}     & \multicolumn{4}{c|}{ \textbf{FGVC-Aircraft}}   & \multicolumn{3}{c}{\textbf{Stanford Cars}} \\
    \cmidrule{2-12}          & \textbf{order\_acc} & \textbf{family\_acc} & \textbf{specie\_acc} & \textbf{avg\_acc}   & \textbf{maker\_acc} & \textbf{family\_acc} & \textbf{model\_acc} & \textbf{avg\_acc}   & \textbf{maker\_acc} & \textbf{model\_acc }& \textbf{avg\_acc}   \\
    \midrule
    \midrule
    \textbf{Vanilla\_single} & $95.38\pm0.47$  & $87.70\pm0.79$ & $74.24\pm0.86$ & $85.77$         & $90.82\pm1.02$ &$ 88.73\pm1.17$ & $86.26\pm1.37$ & $88.60 $         & $95.30\pm0.11$ & $88.66\pm0.45$ & $91.98$  \\
    \textbf{Vanilla\_multi}  & $95.13\pm0.53$  & $89.70\pm0.13$ & $78.31\pm0.35$ & $87.71$         & $90.69\pm0.48$ & $89.23\pm0.53$ & $88.10\pm0.10$  & $89.34$         & $95.24\pm0.20$ & $89.14\pm0.16$ & $92.19$  \\
    \textbf{Ours\_single}  & $95.63\pm0.27$  & $88.50\pm0.15$ & $77.46\pm0.10$ & $87.50$         & $90.73\pm0.23$ & $89.39\pm0.11$ & $87.96\pm0.27$ & $89.36 $         & $95.23\pm0.09$ & $89.12\pm0.29$ & $92.18 $  \\
    \textbf{Ours}            & $96.37\pm0.16$  & $90.39\pm0.15$ & $77.95\pm0.04$ & $88.24$         & $\textbf{\blue{93.04$\pm$0.25}}$ & $90.73\pm0.19$ & $\textbf{\blue{88.35$\pm$0.18}}$ & $\textbf{\blue{90.71}}$          & $95.58\pm0.06$ & $89.66\pm0.16$ & $92.62$  \\
    \midrule
    \textbf{Ours\_MC}        &$96.58\pm0.15$ & $90.36\pm0.07$ & $77.85\pm0.38 $& $88.26 $ &$ 92.86\pm0.12$ &$ 90.74\pm0.11$ & $88.19\pm0.11$ &$ 90.59$  & $95.56\pm0.17$ & $89.62\pm0.21$ & $92.59$ \\
    \midrule
    \textbf{Ours\_NTS}       & $\textbf{\blue{96.57$\pm$0.07}}$  & $\textbf{\blue{91.58$\pm$0.57}}$ & $\textbf{\blue{80.45$\pm$0.68}}$ & $\textbf{\blue{89.53}}$         & $92.48\pm0.16$ & $\textbf{\blue{90.75$\pm$0.07}}$ & $88.31\pm0.23$ & $90.51$          & $\textbf{\blue{95.96$\pm$0.39}}$ & $\textbf{\blue{90.64$\pm$0.37}}$ & $\textbf{\blue{93.30}}$  \\
    \midrule
    \textbf{Ours\_PMG}       & $\textbf{\red{97.98$\pm$0.12}}$  & $\textbf{\red{93.50$\pm$0.10}}$ & $\textbf{\red{82.26$\pm$0.13}}$ & $\textbf{\red{91.25}}$& $\textbf{\red{94.57$\pm$0.10}}$ & $\textbf{\red{92.44$\pm$0.07}}$ & $\textbf{\red{89.62$\pm$0.15}}$ & $\textbf{\red{92.21}}$ &
    $\textbf{\red{96.42$\pm$0.05}}$              &$\textbf{\red{91.05$\pm$0.15}}$              &$\textbf{\red{93.74}}$  \\
    \bottomrule
    \end{tabular}%
  \end{adjustbox}
    \caption{Comparisons with different baselines for FGVC task under multi-granularity label setting.}
  \label{tab:results}%

\end{table*}%

\section{Experimental Settings}

\keypoint{Datasets} We evaluate our proposed method on three widely used FGVC datasets. While some dataset only offers one fine-grained label for each of its images, we manually construct a taxonomy of label hierarchy by tracing their parent nodes (superclasses) in Wikipedia pages. Details are as follows. (i) \textbf{CUB-$\bf{200}$-$\bf{2011}$}~\cite{wah2011caltech} is a dataset that contains $11,877$ images belonging to $200$ bird species. We re-organise this dataset into three-level label hierarchy with $13$ orders (e.g., ``Passeriformes'' and ``Anseriformes''), $38$ families (e.g.,  ``Icteridae'' and ``Cardinalidae'' ) and $200$ species (e.g., ``Brewer Blackbird''  and ``Red winged Blackbird''). (ii) \textbf{FGVC-Aircraft}~\cite{maji13fine} is an aircraft dataset with $10,000$ images covering $100$ model variants. It comes with three-level label hierarchy with $30$ makers (e.g., ``Boeing'' and ``Douglas Aircraft Company''), $70$ families (e.g.,`` Boeing $767$'',`` Boeing $777$''), and $100$ models (e.g., ``$767$-$200$'', ``$767$-$300$''), which we directly adopt for our setting. (iii) \textbf{Stanford Cars}~\cite{krause20133d} contains $8,144$ car images categorised by $196$ car makers.  We re-organise this dataset into two-level label hierarchy with $9$ car types (e.g., ``Cab'' and ``SUV'') and $196$ specific models (e.g., ``Cadillac Escalade EXT Crew Cab $2007$'' and ``Chevrolet Avalanche Crew Cab $2012$''). We follow the standard train/test splits as laid out in the original datasets. We do not use any bounding box/part annotations in all our experiments.

\keypoint{Implementation details} For fair comparisons, we adopted ResNet$50$ pre-trained on ImageNet as our network backbone and resize each input image to $224\times224$ throughout the experiments unless otherwise specified. We set the number of hidden units in $f$ as $512$ when a single model is asked to predict one label only, and $600$ when that is adapted for multiple labels. To deal with the imbalance between ImageNet pre-trained convolutional layers and newly added fully-connected layers in the classification heads, we adopt different learning rates starting from $0.01$ and $0.1$ respectively. Common training augmentation approaches including horizontal flipping and random cropping, as well as colour jittering are applied. We train every single experiment for $100$ epochs with weight decay value as $5\times10^{-4}$. \texttt{MomentumOptimizer} is used with momentum value $0.9$ throughout. The code will be made publicly accessible.

\keypoint{Evaluation metrics} Following community convention, FGVC performance is quantified by acc, the percentage of images whose labels are correctly classified. We use avg\_acc to calculate the mean of the performance across label granularities. Each experiment is run three times. The mean and standard deviation of the results obtained over three trials are then reported.

\keypoint{Baselines} As our focus is on how to adapt an image classification model with single label output into multiple ones, our baselines comprise alternative multi-label classification models. To show our proposed solution is generic to any existing FGVC frameworks, we also include three other baselines by replacing the backbone of our model with different advanced FGVC-specific components. \textbf{Vanilla\_single}: this corresponds to one single shared network backbone with multiple classification heads appended to the end. \textbf{Vanilla\_multi} adopts one independent network backbone for each label prediction. \textbf{Ours\_single} improves upon Vanilla\_single aiming to disentangle the decision space in multi-granularity label predictions. This is achieved by splitting $f$ into equal number of segments as that of classifiers, with each independently responsible for one classifier at one granularity. \textbf{Ours} advances Ours\_single in better feature disentanglement by reinforcing coarse-grained classifiers with fine-grained features. Finally, \textbf{Ours\_MC}~\cite{chang2020mc}, \textbf{Ours\_NTS}~\cite{yang2018learning}, \textbf{Ours\_PMG}~\cite{du2020fine}, represent three means of training our proposed method on top of state-of-the-art FGVC frameworks.

\section{Results and Analysis}

\subsection{Comparison with Baselines}

Our experimental discovery coincides well with our intuition that compared with classifying one fine-grained label, there exists additional issue that needs to be taken care of in multi-granularity label predictions. Our proposed method can not only effectively solve this problem, but also generic in terms of the network backbone used. Belows is more detailed analysis of the results with reference to Table~\ref{tab:results}.

\keypoint{\emph{Is our model effective in solving FGVC problem with multi-granularity label output?}} Yes. It is evident that the proposed model (Ours) outperforms all other baselines under the metric of avg\_acc on all three datasets. Furthermore, the consistent performance gain from Our\_MC to Ours\_NTS, and to Ours\_PMG tells one important message: our solution not only supports easy drop-in to existing FGVC models, but also does not undermine their original functionality when adapted.

\keypoint{\emph{Are the proposed technical contributions appropriate?}} Yes. The significant gap between Vanilla\_single and Ours\_single confirms the severity of feature entanglement between label granularities - that can be alleviated by simply splitting a feature into several parts with each corresponding to an independent classifier. The proposed Reinforce module (Ours\_single vs. Ours) is effective in boosting the 
classification performance at coarse granularity (e.g., order\_acc and family\_acc in CUB-$200$-$2011$). The fact that it can also achieve higher accuracy on the finest labels (e.g., species\_acc), a task which has not been explicitly designed to improve on, provides direct evidence of how better feature disentanglement is further taking place.

\keypoint{\emph{What does Vanilla\_multi tell us?}} The performance of Vanilla\_multi draws our attention. On one hand, its accuracy on the finest label prediction crushes all opponents by significant margins across the datasets. On the other, it performs the worst on classifying coarsest labels. Such contrast, however, echoes our observation that underlies the technical considerations of this paper: finer-grained classifier performs the best when it is portrayed as a single independent task itself, while coarser-level label predictions can benefit significantly from a multi-granularity task setting. Note that since Vanilla\_multi requires equal number of unshared network backbones as that for classification tasks, it is not a strictly fair comparison in terms of its model capacity. The purpose here is to show solving disentanglement between label prediction at different granularities remains challenging, albeit we have greatly advanced the problem.

\begin{figure*}[t]
\begin{center}
  \includegraphics[width=0.98\linewidth]{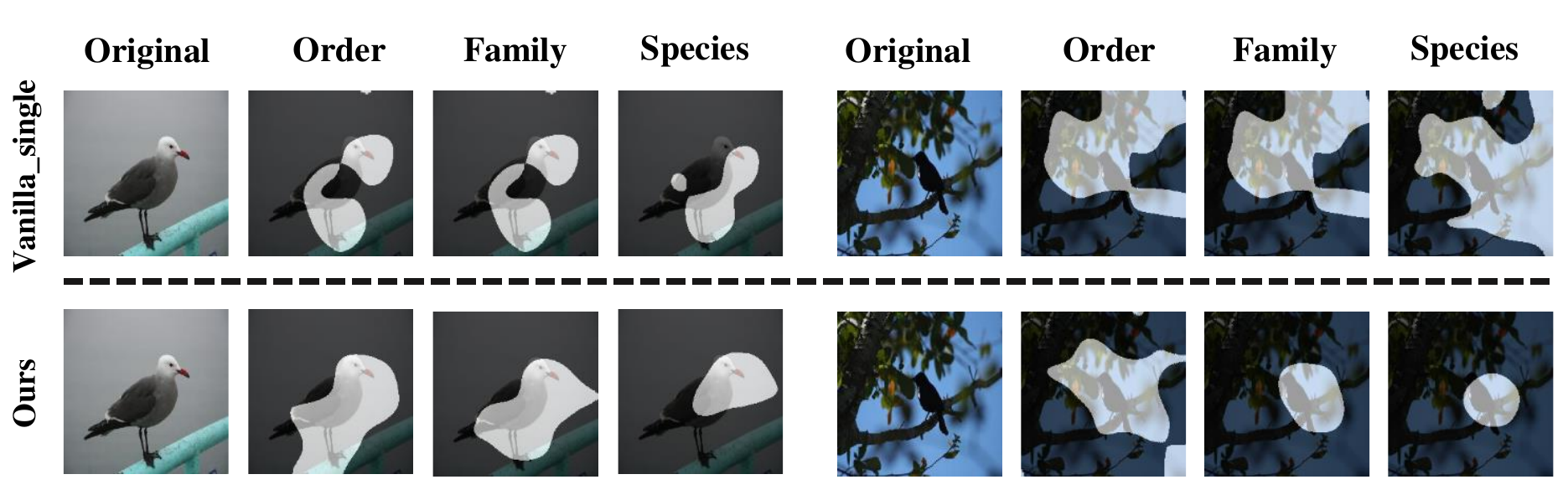}
\end{center}

  \caption{We highlight the supporting visual regions for classifiers at different granularity of two compared models. Order, Family, Species represent three coarse-to-fine classifiers trained on CUB-$200$-$2011$ bird dataset.}
\label{fig:vis}
\vspace{-5mm}
\end{figure*}

\keypoint{\emph{What does it look like?}} {We further carry out model visualisation to demonstrate that classifiers [$\mathcal{G}_1$, ..., $\mathcal{G}_K$] under Vanilla\_single and Ours indeed capture different regions of interests that are useful for FGVC, and offer insight on how better disentanglement is taking place. To this end, we adopt Grad-Cam \cite{selvaraju2017grad} to visualise the different image supports for each $\mathcal{G}_k$ by propagating their gradients back to $x$. It can be seen from the bottom half of Figure \ref{fig:vis} that our classifiers at different hierarchy levels attends to different scales of visual regions -- a clear sign of the model's awareness on coarse-fine disentanglement. In contrast, the top half of Figure \ref{fig:vis} shows that Vanilla\_single appears to focus on similar un-regularised image parts across label granularity.} 

\begin{table}[t]
  \centering
  \footnotesize

\begin{adjustbox}{width=\linewidth,center}
    \begin{tabular}{l@{}c@{}c@{}c}
    \hline
    \textbf{Method   }                                      & \textbf{CUB-200-2011}      \        &\textbf{ FGVC-Aircraft }                &\textbf{ Stanford Cars }   \\
    \hline
    \hline
    \textbf{FT ResNet}~\cite{wang2018learning}                  & $84.1$                                            & $88.5$                        & $91.7$            \\
    \textbf{DFL}~\cite{wang2018learning}                    & $87.4$                                            & $91.7$                        & $93.1$             \\
    \textbf{NTS}~\cite{yang2018learning}                    & $87.5$                                            & $91.4$                        & $93.9$   \\
    \textbf{TASN}~\cite{Zheng_2019_CVPR}                      & $87.9$                                            &      -                        & $93.8$                  \\
    \textbf{API-Net}~\cite{zhuang2020learning}                  & $87.7$                                            & $93.0$                        & $\textbf{\blue{94.8}}$   \\    
    \textbf{MC-Loss}~\cite{chang2020mc}                         & $87.3$                                            & $92.6$                        & $93.7$  \\   
    \textbf{PMG}~\cite{du2020fine}                             & $\textbf{\blue{89.6}}$                       & $\textbf{\blue{93.4}}$   & $\textbf{\red{95.1}}$   \\   
    \hline
    \hline

    \textbf{Ours  }                                                      & $86.8$                                              & $92.8$                         &  $94.3$  \\   
    \textbf{Ours\_PMG }                                                  & $\textbf{\red{89.9}}$                               & $\textbf{\red{93.6}}$          &  $\textbf{\red{95.1}}$       \\  
    \hline
    \hline
    \textbf{Ours\_HC }                                                  & $85.0$            & $90.6$                & $92.8$       \\   
    \textbf{Ours\_DFT }                                                 & $85.5$            &  $91.7$             &  $93.2$     \\   
    \hline
    \end{tabular}%
\end{adjustbox}
  \caption{Performance comparisons on traditional FGVC setting with single fine-grained label output.}
  \label{tab:SOTA}%
\vspace{-6mm}
\end{table}%

\subsection{Evaluation on traditional FGVC setting}
Our model can be evaluated for FGVC without any changes -- we just need to report classification accuracy for fine-grained labels at the bottom of the hierarchy. However, for fair comparison with other state-of-the-art FGVC works, we also resize image input to a size of $448$ $\times$ $448$. We leave all other implementation settings unchanged, and do not perform grid search for performance gain. The results are reported in Table~\ref{tab:SOTA}. We can see that by building our method upon the backbone of PMG, new state-of-the-art results (Ours\_PMG) for traditional FGVC setting are gained on CUB-$200$-$2011$ and FGVC-Aircraft datasets. Improvements over state-of-the-art on Stanford Cars dataset is less significant. We attribute this to the relatively shallow hierarchy (two levels) on Stanford Cars. Note that we do not introduce any extra parameters when implemented on top of traditional FGVC methods. 

\keypoint{The role of label hierarchy}  To investigate the impact of label hierarchy on the traditional FGVC performance, we compare our manual method of constructing label hierarchy based on Wikipedia pages with two variants, Hierarchical Clustering (Ours\_HC) and Deep Fuzzy Tree (Ours\_DFT)~\cite{wang2019deep}. These are two clustering methods that automatically mine hierarchical structures from data, which mainly differ in how to measure the distance between clusters and whether there are tree structures explicitly modelled. For both methods, we stop the discovery process when three-level label hierarchy has been formed. From the last two rows in Table~\ref{tab:SOTA}, the following observations can be made: (i) Manual hierarchies achieves the best performance across all three datasets, suggesting semantically defined parent-child relationships tend to encourage cross granularity information change. (Ours vs. Ours\_HC vs. Ours\_DFT); (ii) Traditional FGVC problem (FT ResNet) benefits from multi-granularity label setting, regardless of what label hierarchy is used.

\section{Discussions}
Here, we offer discussions on some potentially viable future research directions, with the hope to encourage follow up research. 

\keypoint{Beyond multi-task learning} While our MTL framework has shown promise as a first stab, other means of encouraging information exchange/fusion across hierarchy levels can be explored. One possible alternative is via meta learning~\cite{hospedales2020meta}. In this sense, rather than learning multi-granularity label prediction task in one shot, we can treat them as a sequence of related tasks optimised over multiple learning episodes. An idea could be that in the inner loop, we find a meta-learner that serves as good initialisation with few gradients away to each task (as per disentanglement). We then ask the outer task-specific learners to quickly adapt from it (as per reinforcement). 

\keypoint{From classification to retrieval.} Formulating the problem of fine-grained visual analysis as a classification task itself underlies certain limitations: the fixed number of labels makes it rigid to be applied in some open-world scenarios {\cite{wei2019deep}}. By projecting images into a common embedding space (as per retrieval) however, we will not only grant the flexibility but also potentially relax the ways of granularity interpretation into model design. Pretending that we were to address the goal of this paper from a retrieval perspective, we can associate label granularity with the model's receptive field -- the finer the label, the more local the regions of interest. We can also potentially directly use label granularity as an external knowledge to dynamically parameterise the embedding space (as per hypernetworks ~\cite{ha2016hypernetworks}). More importantly, a successfully-trained model now has a chance to learn a smooth interpolation between label granularities, which is of great practical value but infeasible under the formulation of classifiers. 

\keypoint{Rethinking ImageNet pre-training} FGVC datasets remain significantly smaller than modern counterparts on generic classification {\cite{deng2009imagenet,peng2019moment}}. This is a direct result of the bottleneck on acquiring expert labels. Consequently, almost all contemporary competitive FGVC models rely heavily on pre-training: the model must be fine-tuned upon the pre-trained weights of an ImageNet classifier. While useful in ameliorating the otherwise fatal lack of data, such practice comes with a cost of potential mismatch to the FGVC task -- model capacity for distinguishing between ``dog''' and ``cat'' is of little relevance with that for differentiating ``\fred{Giant Ibis}'' and ``\fred{flamingo}''. In fact, our paper argues otherwise -- that coarse-level feature learning is best disentangled from that of fine-grained. Recent advances on self-supervised representation learning provide a promising label-efficient way to tailor pre-training approaches for downstream tasks {\cite{pang2020solving,tschannen2020self}}. However, its efficacy remains unknown for FGVC.

\section{Conclusion}
Following a human study, we re-envisaged the problem of fine-grained visual classification, from the conventional single label output setting, to that of coarse-fine multi-granularity label prediction. We discovered important insights on how positive information exchange across granularities can be explored. We then designed a rather simple yet very effective solution following these insights. Extensive experiments on three challenging FGVC datasets validate the efficacy of our approach. When evaluated on the traditional FGVC setting, we also report state-of-the-art results while not introducing any extra parameters. We will release all human study data, and make our code publicly accessible. Last but not least, we hope to have caused a stir, and trigger potential discussions on the very title of this paper -- that whether my ``Flamingo'' should or should not be your ``Bird''.

{\small
\bibliographystyle{ieee_fullname}
\bibliography{main}
}

\typeout{get arXiv to do 4 passes: Label(s) may have changed. Rerun}
\end{document}